\documentclass{article}

\usepackage[preprint]{corl_2026} 
\usepackage{graphicx}
\usepackage{amsfonts}
\usepackage{amsmath}
\usepackage{subcaption}
\usepackage{multirow}
\usepackage{enumitem}
\usepackage{fix-cm}   
\usepackage{graphicx}

\newcommand{\smalls}{\fontsize{8}{9}\selectfont}
\title{Dreaming when Necessary: Advancing World Action Models with Adaptive Multi-Modal Reasoning}

\author{
  Yinzhou Tang$^*$\\
  Tsinghua University \\
  \And
  Jingbo Xu$^*$ \\
  Tsinghua University \\
  \AND
  Yu Shang$^*$ \\
  Tsinghua University \\
  \And
  Zihao Song \\
  Tsinghua University \\
  \And
  Chen Gao \\
  Tsinghua University \\
  \And
  Wei Wu \\
  Manifold AI \\
  \And
  Yong Li \\
  Tsinghua University \\
  \vspace{0.3em}\\
  \footnotesize{$^*$Equal Contribution}
}

\begin{document}
\maketitle


\begin{abstract}
World Action Models (WAMs) offer a promising approach to embodied intelligence, yet existing methods rely heavily on video prediction as action priors and lack adaptive multimodal reasoning, limiting their effectiveness on long-horizon, complex tasks. We observe that WAMs require different multimodal reasoning modes under different execution contexts: textual reasoning is essential during task transitions to guide high-level action prediction, while visual reasoning is critical during fine-grained manipulation for precise control. Motivated by this observation, we propose \textbf{AdaWAM}, a world action model with adaptive multimodal reasoning abilities. AdaWAM integrates a lightweight dynamic router that autonomously triggers textual or visual reasoning as needed during task execution. Experiments on both simulated and real-world embodied tasks show that AdaWAM substantially improves inference efficiency while outperforming state-of-the-art embodied policies. Codes and demos are available at: \url{https://adawam.github.io/}.
\end{abstract}



\section{Introduction}

Recent progress in World Action Models (WAMs) has introduced a promising paradigm for embodied intelligence~\cite{wang2026world,ye2026world}. Unlike conventional Vision-Language-Action (VLA) models~\cite{sapkota2025vision,ma2026survey,zhang2025pure}, WAMs explicitly model the coupling between future observations and robot actions, enabling policies to reason about physical dynamics rather than merely react to current visual states. By leveraging video prediction as an action prior, WAMs can simulate possible future rollouts and improve action generation in contact-rich and physically complex scenarios~\cite{ye2026world,li2026causal,yuan2026fastwam}.

Existing WAMs mainly follow two paradigms. The first is \emph{video-action joint prediction}, where future visual observations and actions are generated together to provide physical foresight for policy execution~\cite{ye2026world,bi2025motus,li2026causal}. This paradigm improves action reliability, especially during fine-grained manipulation, but incurs substantial computational latency due to co-decoding video and action tokens. The second is \emph{action-only prediction}, which directly decodes actions to improve inference efficiency~\cite{yuan2026fastwam}. However, without explicit visual foresight, action-only models often behave reactively and struggle at critical manipulation steps that require precise physical anticipation. Moreover, most existing WAMs use language merely as a static task description, leaving rich textual semantics underexplored for long-horizon task understanding and stage-wise action guidance.

We argue that the core limitation of current WAMs lies in their lack of adaptive multimodal reasoning. Embodied task execution requires different reasoning modes at different stages: textual reasoning is crucial during task transitions for subtask recognition and high-level action guidance, while visual reasoning is essential during fine-grained manipulation, such as grasping, insertion, and alignment, for physical anticipation and precise control. In contrast, many intermediate motion steps can be handled efficiently through action-only decoding. Our controlled studies further support this observation: removing text during task-transition stages leads to severe performance degradation, while video-action joint prediction mainly outperforms action-only prediction during fine-grained manipulation, with both paradigms performing similarly during simple motion.

\begin{figure}[t]
    \centering
    \includegraphics[width=0.99\linewidth]{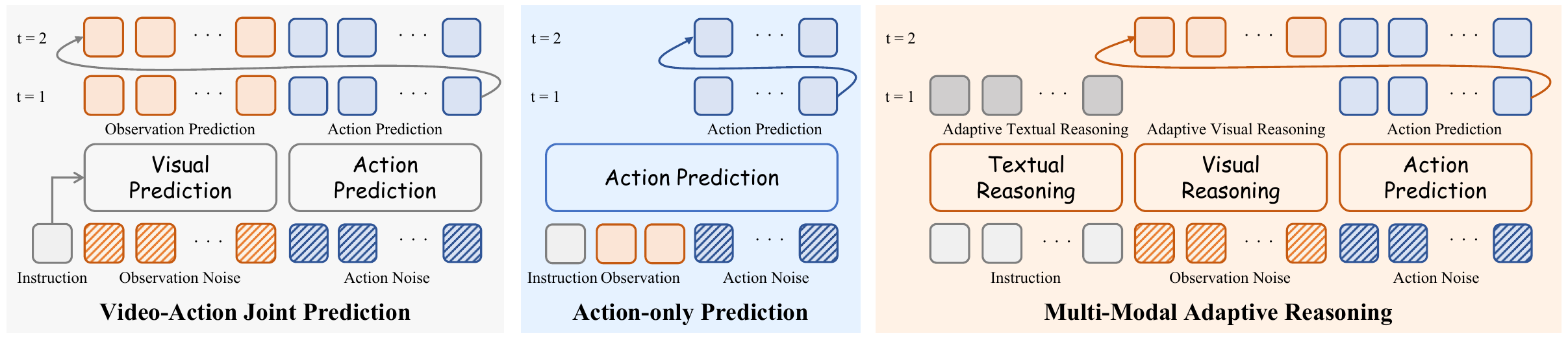}
    \caption{Paradigm comparison among video-action joint prediction, action-only prediction, and the proposed adaptive multimodal reasoning framework.}
    \label{fig:intro}
    \vspace{-15pt}
\end{figure}

Motivated by these findings, we propose \textbf{AdaWAM}, a world action model with adaptive multimodal reasoning. To support adaptive reasoning, we first build a scalable multimodal reasoning annotation pipeline. For long-horizon demonstrations, the pipeline parses trajectory cues, such as end-effector motion, gripper states, and task-relevant motion patterns, to locate subtask transitions and fine-grained manipulation intervals. A VLM then verifies visual observations within these trajectory-guided windows, producing dense per-frame labels for language-conditioned subtask boundaries and precision-critical manipulation phases. These annotations supervise when textual or visual reasoning should be activated.
In terms of model design, AdaWAM introduces a lightweight dynamic router that autonomously selects the appropriate reasoning mode during task execution. Specifically, it activates textual reasoning when high-level task understanding is required, triggers visual reasoning when physical foresight is needed for fine-grained control, and otherwise defaults to efficient action-only decoding. This forms an interleaved multimodal reasoning chain, enabling AdaWAM to adaptively choose the most suitable reasoning mode for action prediction under different execution contexts. As a result, AdaWAM preserves the benefits of multimodal reasoning while avoiding unnecessary inference overhead.
Extensive evaluations on LIBERO~\cite{liu2023libero}, RoboTwin 2.0~\cite{chen2025robotwin}, and real-world ALOHA/PiPER tasks show that AdaWAM consistently improves performance on long-horizon and fine-grained manipulation tasks. In particular, AdaWAM achieves the best results on LIBERO-Long and the hard subsets of RoboTwin 2.0, while also demonstrating stronger real-world task competence than prior WAM and VLA methods.

Our main contributions are summarized as follows:
\begin{itemize}[leftmargin=*]

    \item We propose AdaWAM, a World Action Model framework with adaptive multimodal reasoning. Unlike static WAM pipelines, AdaWAM dynamically interleaves textual reasoning, visual reasoning, and action-only decoding according to the current execution context.

    \item We formulate adaptive inference as token-level routing and introduce a dynamic router, which autonomously activates textual or visual reasoning for efficient action prediction.

    \item Extensive experiments on both simulated and real-world embodied tasks demonstrate that AdaWAM substantially improves inference efficiency while outperforming state-of-the-art embodied policies.
\end{itemize}


\section{Related Work}
\subsection{Vision Language Action Models}
Vision Language Action (VLA) models leverage pretrained vision language representations and large-scale robot demonstrations for language-conditioned robot control~\citep{brohan2022rt1, brohan2023rt2, kim2025openvla, black2024pi0, liu2024rdt, bjorck2025gr00t, gemini2025robotics}. Despite strong generalization, direct action prediction in many VLA and generalist policies limits explicit reasoning over task progress, subgoals, and fine object interactions in long-horizon manipulation~\citep{zitkovich2023rt2, kim2025openvla, ghosh2024octo}. Recent work addresses this issue with explicit reasoning supervision, such as textual rationales or visual chain of thought subgoals~\citep{zawalski2024ecot, zhao2025cotvla}. In contrast, our work studies language-guided reasoning in video-based World Action Models (WAMs), where action prediction is coupled with learned visual dynamics and world-grounded representations.
\vspace{-10pt}
\subsection{World Action Models}
\vspace{-5pt}
A parallel direction studies robot control through video-based world modeling, where models learn how scenes evolve under interaction by predicting future visual states, trajectories, or action-conditioned videos~\citep{du2023learning, zhou2024robodreamer, bharadhwaj2024gen2act, cheang2024gr2, hu2024vpp}. Recent World Action Models (WAMs) extend this idea by jointly modeling future video and robot actions in unified generative frameworks~\citep{zhu2025unified, pai2025mimicvideo, kim2026cosmos, li2026causal, ye2026world}. DreamZero uses pretrained video diffusion backbones for joint video action prediction, while Fast-WAM shows that WAM benefits largely arise from video cotraining rather than costly future generation at inference~\citep{ye2026world,yuan2026fastwam}. Although language and visual intermediates improve action grounding~\citep{zawalski2024robotic, zhao2025cotvla}, their role in deciding when to invoke test-time imagination remains unclear. Our work addresses this gap by using adaptive multimodal reasoning to balance efficient action-only inference with robust joint video action prediction.

\vspace{-5pt}
\section{Method}
\vspace{-5pt}
\subsection{Multimodal Reasoning Data Annotation Pipeline}
\vspace{-5pt}
\begin{figure}[t]
    \centering
    \includegraphics[width=0.95\linewidth]{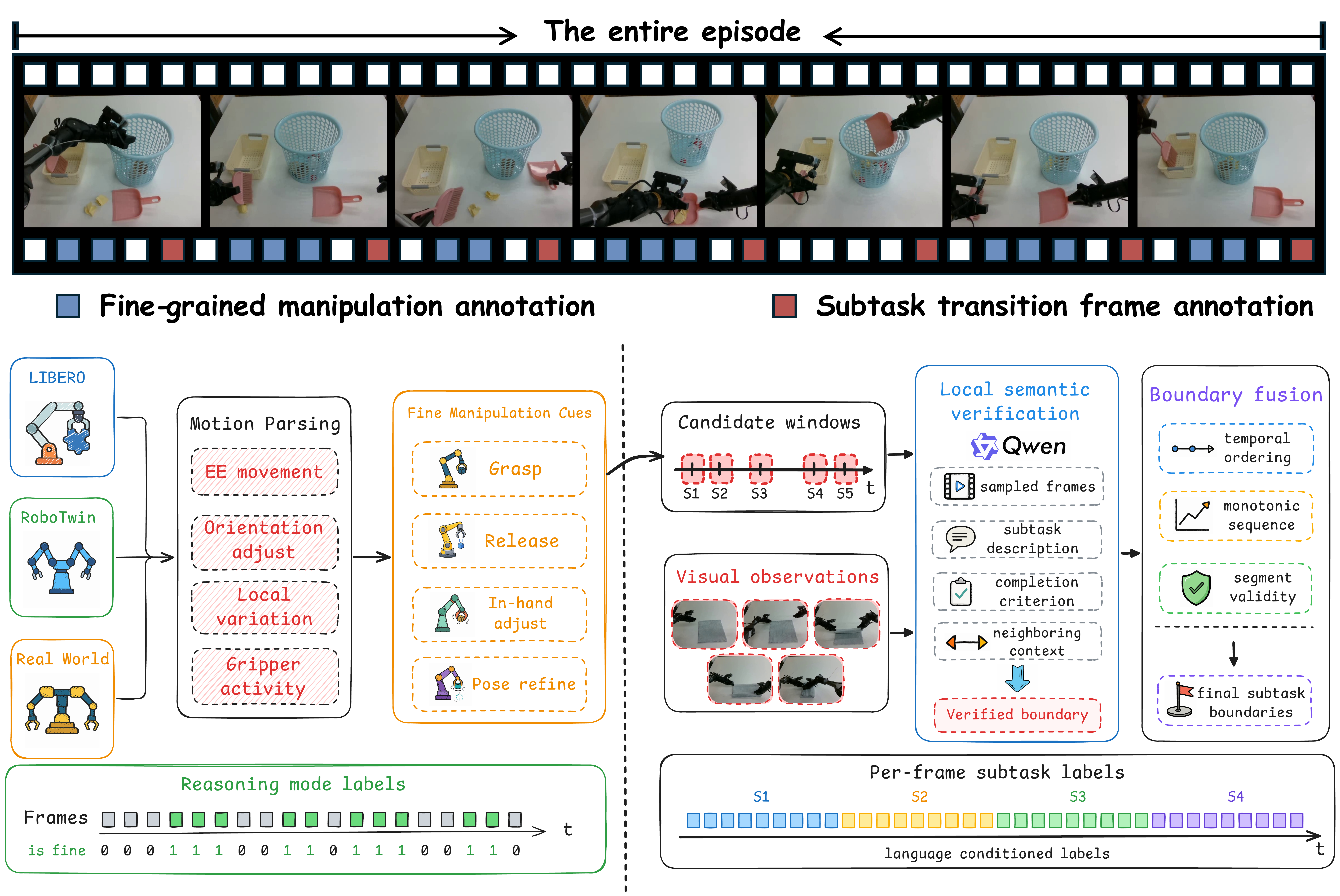}
    \vspace{-5pt}
    \caption{Overview of the annotation pipeline. Trajectory cues localize fine manipulation intervals and candidate subtask windows, while VLM verification produces subtask labels.
}
    \label{fig:data-pipeline}
    \vspace{-15pt}
\end{figure}
\paragraph{Trajectory-guided Subtask Annotation.}
To annotate subtask boundaries in long-horizon manipulation demonstrations, we develop a hybrid alignment pipeline. For each demonstration, robot state trajectories are parsed to extract physically grounded cues, including end-effector motion, gripper transitions, and task-relevant motion patterns. These cues generate candidate temporal windows for each predefined subtask, reducing global video segmentation to local verification around meaningful physical events. Within each candidate window, we use Qwen3-VL 8B~\cite{Qwen3-VL} as a semantic verifier to identify subtask completion. Given sampled visual observations, the subtask description, completion criterion, and neighboring context, the model predicts the earliest frame where the current subtask is visually and stably completed. The verified boundaries are fused under temporal ordering and segment validity constraints to enforce a monotonic subtask sequence.

\paragraph{Motion-Based Fine Manipulation Labeling.}
To provide fine-grained action supervision, we annotate critical manipulation intervals across LIBERO~\cite{liu2023libero}, RoboTwin~\cite{chen2025robotwin}, and our real-world experiments using a unified trajectory-based procedure. Robot states are converted into motion patterns that capture end-effector displacement, orientation adjustment, local motion variation, and gripper activity. These signals distinguish fine manipulation phases, such as grasping, releasing, in-hand adjustment, and local pose refinement, from coarse reaching or repositioning. We then temporally refine the detected intervals to remove isolated noisy frames and retain continuous segments.
\vspace{-10pt}
\subsection{Model Architecture}
\vspace{-10pt}
\begin{figure}[t]
    \centering
    \includegraphics[width=0.95\linewidth]{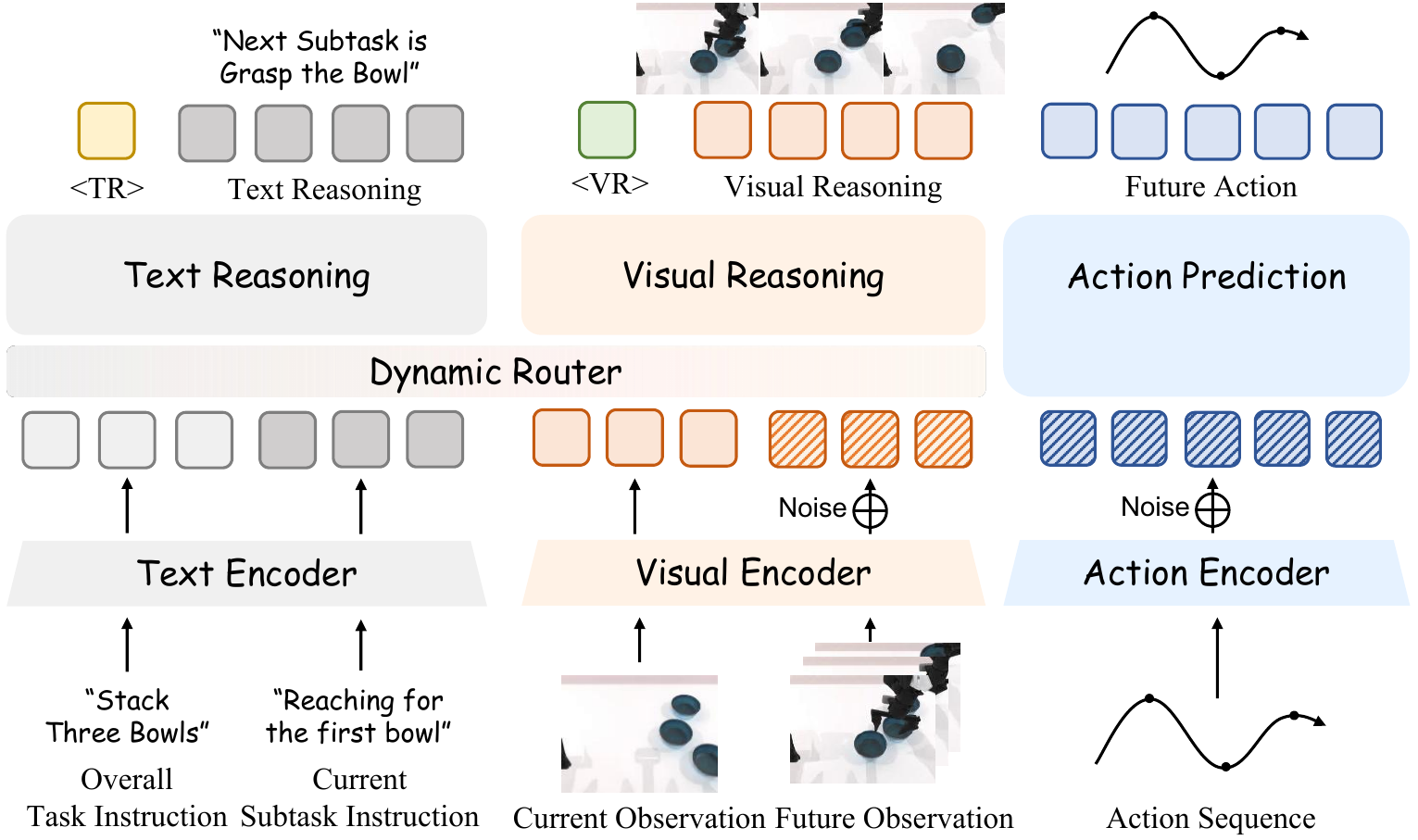}
    \caption{Model architecture for AdaWAM. It is able to conduct adaptive multi-modal reasoning with the text reasoning module and the visual reasoning module controlled by the dynamic router, and predict future actions with the action prediction module.}
    \label{fig:main}
    \vspace{-15pt}
\end{figure}
We formulate robotic manipulation as a conditional generation problem within a latent space. Let $o_t \in \mathbb{R}^{H \times W \times 3}$ be the visual observation, $l$ be the global task instruction, and $c_t$ be the localized subtask text. Using a pre-trained visual encoder $\mathcal{E}$, we map observations into latent tokens $z_t = \mathcal{E}(o_t)$. 
The architecture of \textbf{AdaWAM} achieves adaptive multi-modal reasoning and is constructed upon three core components, including a generative diffusion backbone, a semantic specialist, and an adaptive routing mechanism.
\vspace{-5pt}
\paragraph{Video-Action DiT for Video Reasoning and Action Generation.}
Our foundational generative backbone employs a Diffusion Transformer (DiT) paradigm that for physical dynamics modeling and policy execution. Specifically, the VideoDiT ($\mathcal{M}_\theta$) acts as a latent world model to capture environment dynamics, reversing a forward diffusion process to predict future visual latents $z_{t+1:t+K}$ conditioned on historical states $z_{\leq t}$. Complementary to this, the ActionDiT ($\pi_\phi$) serves as the policy network, generating continuous action chunks $a_{t:t+H}$ by denoising the noise. Crucially, this action generation is conditioned on the current state $z_t$, the subtask embedding $c_t$, and optionally the future physical foresight $z_{t+1:t+K}$, providing dense spatial guidance for contact-rich manipulations.
\vspace{-5pt}
\paragraph{Text Reasoning Module for Text Reasoning.}
To transcend the limitations of static task descriptions in long-horizon tasks, we introduce a Text Reasoning Module ($\mathcal{V}_\omega$) instantiated as a compact Vision-Language Model (VLM). When activated, $\mathcal{V}_\omega$ auto-regressively predicts the tokenized sequence for the next subtask $c_{t+1}$ conditioned on the visual progression and the global instruction $l$. This dynamic text reasoning ensures that the Video-Action DiT receives temporally aligned and fine-grained linguistic conditioning throughout complex task executions.
\vspace{-5pt}
\paragraph{Dynamic Router for Adaptive Multi-Modal Reasoning.}
To achieve adaptive multi-modal reasoning, we introduce a lightweight Dynamic Router $\mathcal{R}_\psi$. It processes context $C_t = [v_t \,\|\, e_l \,\|\, e_{c_t}]$ to independently predict text reasoning token $<\text{TR}>$ and visual reasoning token $<\text{VR}>$, where $v_t$ is the pooled embedding of $z_t$, and $e_l, e_{c_t}$ are the respective text embeddings for overall task and current subtask instruction. Thus, it is able to predict whether to activate the visual and text reasoning module in the current action chunk.
\vspace{-5pt}
\subsection{Multi-task Training}
\vspace{-5pt}
Training AdaWAM requires aligning the discrete routing decisions with the continuous latent representations of the diffusion backbone. Thus, we utilize a two-stage co-training pipeline.

\textbf{Stage 1: Video-Action DiT and Dynamic Router Training.} We concurrently optimize the VideoDiT, ActionDiT, and the Dynamic Router. The generative models are trained using the continuous-time flow matching objective. We formulate the flow that maps the standard Gaussian noise to the data distribution by regressing the vector fields $v_\theta$ and $v_\phi$:
\begin{align}
    \mathcal{L}_{\text{FM}} = \mathbb{E}_{\tau \sim \mathcal{U}(0,1), z_0, z_1, a_0, a_1} \left[ \|v_\theta(z_\tau, \dots) - (z_1 - z_0)\|^2 + \|v_\phi(a_\tau, \dots) - (a_1 - a_0)\|^2 \right],
\end{align}
where $z_\tau = \tau z_1 + (1 - \tau) z_0$ and $a_\tau = \tau a_1 + (1 - \tau) a_0$, with $\tau \in [0, 1]$ being the time step, $z_0, a_0 \sim \mathcal{N}(0, I)$ the noise samples, and $z_1, a_1$ the ground-truth data representations. Simultaneously, the router is supervised via binary cross-entropy ($\mathcal{L}_{\text{BCE}}$) against heuristically annotated routing labels $\hat{y}$ (derived from task phase transitions and contact-rich state metrics). The joint objective is defined as:
\begin{align}
    \mathcal{L} = \mathcal{L}_{\text{FM}} + \lambda \sum_{k \in \{\text{text, video}\}} \mathcal{L}_{\text{BCE}}(y_k, \hat{y}_k).
\end{align}
\textbf{Stage 2: Text Reasoning Module Fine-Tuning.} 
To empower the framework with advanced semantic reasoning without catastrophic forgetting of the physical priors and flow trajectories learned in Stage 1, we freeze the generative backbone and the dynamic router, solely optimizing the text reasoning module. The VLM minimizes the negative log-likelihood of the ground-truth subtask tokens $\mathcal{L}_{\text{NLL}}(c_{t+1} | z_{\leq t}, l)$, strictly on trajectory segments marked by explicit subtask transitions.
\subsection{Test-time Inference with Adaptive Multi-modal Reasoning}
During test-time deployment, AdaWAM dynamically reconfigures its inference pattern for each action chunk based on the discrete tokens $<\text{TR}>$ and $<\text{VR}>$ predicted by the dynamic router.

For each action chunk, we first predict the visual and text reasoning token with the dynamic router, which determines the inference pattern in this chunk. If <TR>=1, the Text Reasoning Module is invoked to auto-regressively update the subtask instruction to the current subtask instruction $\tilde{c}_t$; otherwise, the previous instruction $c_t$ is retained:
\begin{align}
    \tilde{c}_t = \begin{cases} 
    \mathcal{V}_\omega(z_{\leq t}, l) & \text{if } <\text{TR}> = 1 \\ 
    c_t & \text{if } <\text{TR}> = 0 
    \end{cases}
\end{align}
Afterwards, the visual routing token controls the engagement of the world model. If $<\text{VR}> = 1$, the Visual Reasoning Module is activated to synthesize physical foresights $\tilde{z}_{f}$. Conversely, this heavy visual generation is bypassed to save latency if $<\text{VR}> = 0$:
\begin{align}
    \tilde{z}_{f} = \begin{cases} 
    \mathcal{M}_\theta(z_{\leq t}, \tilde{c}_t) & \text{if } <\text{VR}> = 1 \\ 
    \emptyset & \text{if } <\text{VR}> = 0 
    \end{cases}
\end{align}
Finally, the Action Predictor dynamically assembles its multi-modal conditioning set $\mathcal{S}_t = \{z_{\leq t}, \, \tilde{c}_t, \, \tilde{z}_{f}\}$. The continuous action trajectory $a_{t:t+H}$ is then sampled by integrating the flow conditioned on $\mathcal{S}_t$:
\begin{align}
    a_{t:t+H} \sim \pi_\phi(\cdot \mid \mathcal{S}_t).
\end{align}

\section{Experiments}
\subsection{Benchmarks}
\begin{figure}[htbp]
\vspace{-10pt}
  \centering
{
    \includegraphics[width=0.8\textwidth]{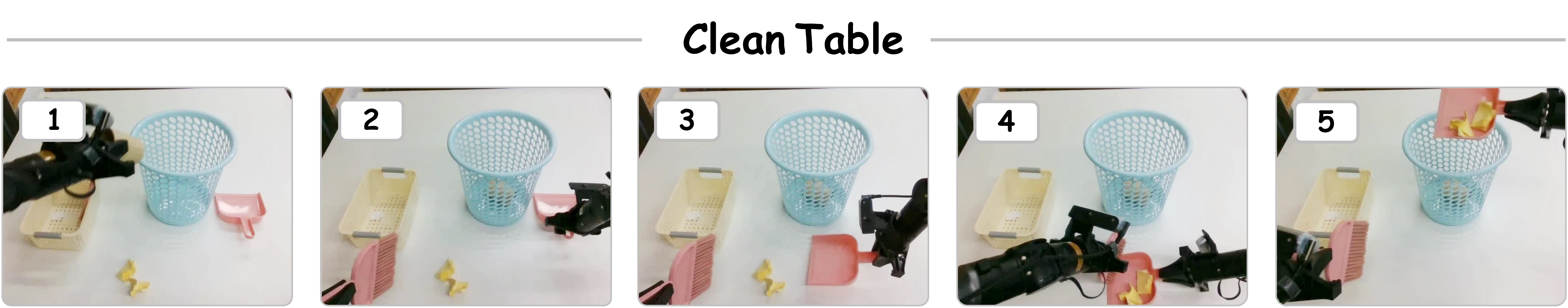}
  }
  \vspace{1pt} 
  \\{
    \includegraphics[width=0.8\textwidth]{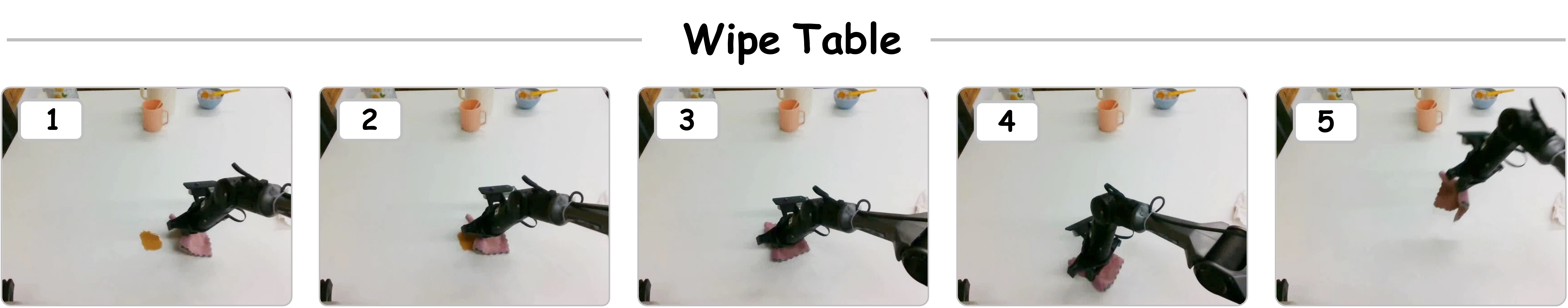}
  }
  \caption{Visualization for real-world long-horizon and fine-grained tasks.}
  \label{fig:real-world-tasks}
  \vspace{-10pt}
\end{figure}

We evaluate our model on three benchmarks from simulation, including LIBERO~\cite{liu2023libero}, RoboTwin2.0~\cite{chen2025robotwin}, and real-world embodiment with AgileX Split-Type ALOHA and PiPER 6-DoF arms, in which we select \textit{Clean Table} and \textit{Wipe Table} as the evaluation task as shown in Figure~\ref{fig:real-world-tasks}. We have also evaluated our model in both trivial tasks and hard tasks subsets with long-horizon operation and fine-grained manipulation. A more detailed description is in Section~\ref{sec:benchmark}. In order to evaluate the adaptive reasoning module, we also evaluate two variants of AdaWAM:
\begin{itemize}[leftmargin=*]
\item \textbf{AdaWAM w/o. T.R.}: In this variant, the textual reasoning module is removed, and it can only use adaptive visual reasoning to achieve video-action joint or action-only prediction.
\item \textbf{AdaWAM w/o. V.R.}: In this variant, we remove the visual reasoning module, and the text reasoning is directly used as the condition for action-only prediciton.
\end{itemize}

\subsection{Evaluation Results}
\paragraph{Overall Performance}
\begin{table}[h]
\vspace{-10pt}
\centering
\smalls
\setlength{\tabcolsep}{15pt} 
\renewcommand{\arraystretch}{1.1} 
\caption{Overall results on LIBERO benchmark.}
\begin{tabular}{l|cccc|c}
\hline
Model            & Spatial & Object & Goal & Long & Overall \\ \hline
OpenVLA          & 84.7    & 88.4   & 79.2 & 53.7 & 76.5    \\
$\pi_0$          & 96.8    & 98.8   & 95.8 & 85.2 & 94.2    \\
CoT-VLA          & 87.5    & 91.6   & 87.6 & 69.0 & 81.1    \\
ACoT-VLA         & \textbf{98.6}    & 99.0   & \textbf{99.4} & 97.0 & \textbf{98.5}    \\
MM-ACT           & 97.8    & 99.4   & 94.8 & 88.0 & 95.0    \\
X-VLA            & 98.2    & 98.6   & \underline{97.8} & 97.6 & \underline{98.1}    \\
LingBot-VA       & \underline{98.5}    & 99.6   & 97.2 & \underline{98.5} & \textbf{98.5}    \\
Motus            & 96.8    & \underline{99.8}   & 96.6 & 97.6 & 97.7    \\
Fast-WAM         & 98.2    & \textbf{100.0}  & 97.0 & 95.2 & 97.6    \\ \hline
AdaWAM w/o. V.R. & 97.5    & 99.4   & 96.8 & 96.6 & 97.6    \\
AdaWAM w/o. T.R. & -       & -      & -    & 97.4 & -       \\
AdaWAM           & 98.0    & 99.6   & 97.1 & \textbf{99.1} & \textbf{98.5}    \\ \hline
\end{tabular}
\label{tab:libero}
\vspace{-10pt}
\end{table}
In order to evaluate the performance of our model in tasks of various difficulties and durations, we utilize LIBERO and RoboTwin 2.0 as the simulation evaluation platform and conduct evaluation in two real-world tasks. In the LIBERO benchmark, we compare our model with both VLA models, such as OpenVLA~\cite{kim2025openvla}, $\pi_0$, X-VLA~\cite{zheng2025x}, and other CoT-based VLAs like CoT-VLA~\cite{zhao2025cot}, ACoT-VLA~\cite{zhong2026acot}, and MM-ACT~\cite{liang2025mm}, and WAMs, including LingBot-VA~\cite{li2026causal}, Motus~\cite{bi2025motus} with video-action joint prediction, and FastWAM~\cite{yuan2026fastwam} with action-only prediction. The results in LIBERO are shown in Table~\ref{tab:libero}. It indicates that AdaWAM achieves the best performance in the LIBERO-Long subset, which consists of long-horizon and multi-object interactions. In the RoboTwin benchmark, we compare our model with other baselines, including GO-1~\cite{bu2025agibot} and $\pi_{0.5}$~\cite{intelligence2025pi_}. The results in RoboTwin .0 are shown in Table~\ref{tab:robotwin}, which indicates that our model achieves the best performance in the selected hard task subset and all tasks in the clean environments. This is owing to the fact that compared to baselines with fine-grained text condition, our model is able to split long-horizon tasks into subtasks, leading to a better performance in long-horizon tasks like \textit{PutObjectCabinet} and \textit{StackBowlsThree}. Furthermore, due to an adaptive visual reasoning paradigm, AdaWAM is able to predict the future action with the guidance of the dreamed future, thus achieving a better performance compared to action-only baselines. Furthermore, we also conduct experiments in the real world to evaluate our model. The results in Table~\ref{tab:real-world} indicate that AdaWAM is also able to execute real-world tasks in both fine-grained and long-horizon tasks.
\begin{table}[]
\centering
\caption{Overall performance on RoboTwin 2.0 benchmark.}
\centering
\setlength{\tabcolsep}{3pt} 
\renewcommand{\arraystretch}{1.1} 
\smalls
\begin{tabular}{lcccccccccc}
\hline
\multicolumn{1}{l|}{Task}               & GO-1  & $\pi_0$   & $\pi_{0.5}$  & X-VLA & LingBot-VA & Motus & Fast-WAM & \begin{tabular}[c]{@{}c@{}}AdaWAM\\ w/o. V.R.\end{tabular} & \begin{tabular}[c]{@{}c@{}}AdaWAM\\ w/o. T.R.\end{tabular} & AdaWAM               \\ \hline
\multicolumn{11}{c}{Clean}                                                                                                                                                                                                                               \\ \hline
\multicolumn{1}{l|}{\textit{HangingMug}}         & 0     & 14    & 18    & 23    & 40    & 38    & \underline{58}    & 56    &  56   &  \textbf{59}   \\
\multicolumn{1}{l|}{\textit{PickDiverseBottles}} & 61    & 69    & 81    & 58    & \underline{89}    & \textbf{90}    & 80    & 79    &  81   &  87   \\
\multicolumn{1}{l|}{\textit{PutObjectCabinet}}   & 60    & 85    & 80    & 46    & 80    & 88    & \underline{94}    & 92    &  91   &  \textbf{96}   \\
\multicolumn{1}{l|}{\textit{RotateQRCode}}       & 22    & 74    & 89    & 34    & \textbf{96}    & 89    & 93    & \underline{95}    &  \textbf{96}   &  94   \\
\multicolumn{1}{l|}{\textit{ScanObject}}         & 1     & 55    & 72    & 14    & \textbf{96}    & 67    & 89    & 91    &  88   &  \underline{92}   \\
\multicolumn{1}{l|}{\textit{StackBowlsThree}}    & 4     & 77    & 77    & 76    & \textbf{100}   & 79    & 80    & 77    &  \underline{82}   &  \textbf{100}   \\
\multicolumn{1}{l|}{\textit{StampSeal}}          & 19    & 46    & 79    & 76    & \textbf{96}    & \underline{93}    & 90    & \underline{93}    &  92   &  91   \\ \hline
\multicolumn{1}{l|}{Hard SR}               & 23.86 & 60.00 & 70.86 & 46.71 & \underline{85.29} & 77.71 & 83.43 & 83.29 & 83.71 & \textbf{88.43} \\
\multicolumn{1}{l|}{Overall SR}            & 37.80 & 65.92 & 82.74 & 72.80  & \underline{92.90}  & 88.66 & 91.88 & 91.31 & 91.88 & \textbf{93.11} \\ \hline
\multicolumn{11}{c}{Random}                                                                                                                                                                                                                              \\ \hline
\multicolumn{1}{l|}{\textit{HangingMug}}         & 0     & 11    & 17    & 27    & 28    & 38    & \textbf{62}    & 54    & 53    & \underline{60}    \\
\multicolumn{1}{l|}{\textit{PickDiverseBottles}} & 56    & 31    & 71    & 36    & 82    & \textbf{91}    & 85    & 83    & 82    & \underline{86}    \\
\multicolumn{1}{l|}{\textit{PutObjectCabinet}}   & 43    & 87    & 79    & 48    & 79    & 71    & \underline{89}    & 87    & 88    & \textbf{91}    \\
\multicolumn{1}{l|}{\textit{RotateQRCode}}       & 9     & 70    & 87    & 33    & \underline{91}    & 73    & 89    & 89    & \textbf{94}    & \textbf{94}    \\
\multicolumn{1}{l|}{\textit{ScanObject}}         & 2     & 42    & 65    & 36    & 91    & 66    & \underline{92}    & \textbf{93}    & 90    & 91    \\
\multicolumn{1}{l|}{\textit{StackBowlsThree}}    & 7     & 75    & 71    & 86    & \textbf{98}    & \underline{87}    & 81    & 78    & 77    & \textbf{98}    \\
\multicolumn{1}{l|}{\textit{StampSeal}}          & 13    & 33    & 55    & 82    & \textbf{97}    & 92    & \underline{94}    & 91    & 89    & 86    \\ \hline
\multicolumn{1}{l|}{Hard SR}               & 18.57 & 49.86 & 63.57 & 49.71 & 80.86 & 74.00 & \underline{84.57} & 82.14 & 81.86 & \textbf{86.57} \\
\multicolumn{1}{l|}{Overall SR}            & 36.24 & 58.40 & 76.76 & 72.84 & \underline{91.50}  & 87.02 & \textbf{91.78} & 90.63 & 90.41 & 91.35 \\ \hline
\end{tabular}
\label{tab:robotwin}
\vspace{-15pt}
\end{table}

\begin{table}[]
\centering
\caption{Overall results on real-world tasks.}
\setlength{\tabcolsep}{2pt} 
\renewcommand{\arraystretch}{1.1} 
\smalls
\begin{tabular}{l|cccccccc}
\hline
Task                    & $\pi_{0.5}$ & X-VLA & Motus & GigaBrain-0 & FastWAM & \begin{tabular}[c]{@{}c@{}}AdaWAM\\ w/o. V.R.\end{tabular} & \begin{tabular}[c]{@{}c@{}}AdaWAM\\ w/o. T.R.\end{tabular} & AdaWAM \\ \hline
Clean Up Trash On Table & 60 & 60 & 50 & 10 & 30 & 30 & 50 & \textbf{70} \\
Wipe Table Clean        & 50 & 20 & 20 & 10 & 50 & 60 & 60 & \textbf{60} \\ \hline
\end{tabular}
\label{tab:real-world}
\vspace{-10pt}
\end{table}

\paragraph{Performance on Fine-grained Tasks}

\begin{figure}[h]
    \centering
    \includegraphics[width=0.99\linewidth]{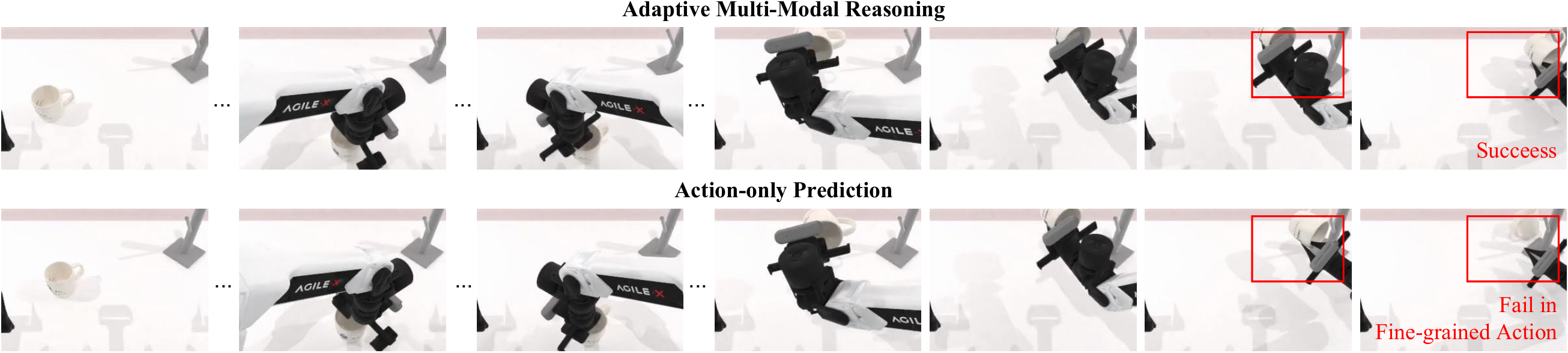}
    \caption{Visualization for action-only WAM and AdaWAM in a fine-grained task (\textit{HangingMug}).}
    \label{fig:hard-tasks}
    \vspace{-10pt}
\end{figure}

We also conduct a case study to evaluate in which cases AdaWAM can outperform the existing baselines. Specifically, for different policies, we set a static seed to ensure the environment is the same for all models. Then we analyze the visualized interaction trajectories. The results in Figure~\ref{fig:hard-tasks} indicate that in fine-grained operations like hanging the mug and grasping the object, our visual reasoning benefits the action prediction with an explicit visual prior.

\subsection{Inference Time Analysis}
\begin{figure}[]
    \centering
    \begin{subfigure}{0.48\linewidth}
        \centering
        \includegraphics[width=0.99\linewidth]{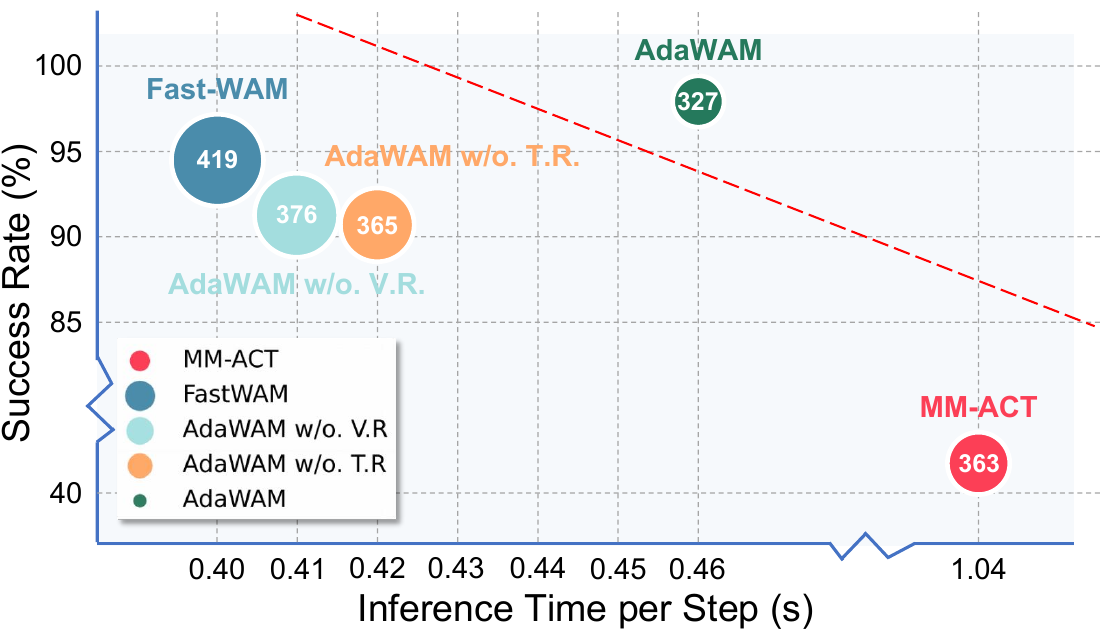}
        \caption{Task \textit{StackThreeBowls}.}
        \label{fig:bowls}
    \end{subfigure}
    \hfill
    \begin{subfigure}{0.48\linewidth}
        \centering
        \includegraphics[width=0.99\linewidth]{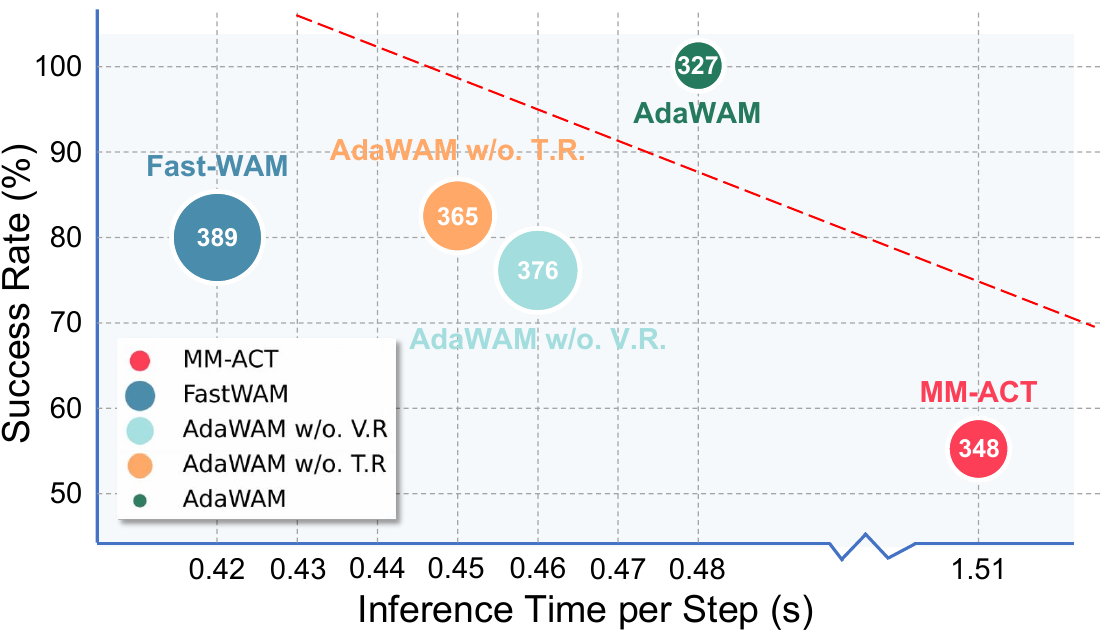}
        \caption{Task \textit{PutObjectCabinet}.}
        \label{fig:cabinet}
    \end{subfigure}
    \vspace{-5pt}
    \caption{Comparison of Inference Time per Step (x-label), Success Rate (y-label), and Duration (size of the bubble).}
    \label{fig:inference-time}
\end{figure}

We also evaluate the inference time compared to other baselines. Specifically, we evaluate three metrics in the \textit{StackThreeBowls} and \textit{PutObjectCabinet} tasks. For each task, we run 100 trajectories and record their success rate, inference time per step, and the total duration to accomplish the task. The results in Figure~\ref{fig:inference-time} indicate that the time efficiency of our model significantly outperforms CoT-based VLAs like MM-ACT. Furthermore, AdaWAM is also comparable to the action-only WAM Fast-WAM but achieves a shorter task duration due to fewer tries with the benefit of adaptive reasoning. For different AdaWAM variants, the inference time of different variants is comparable but faster than that of the original model. This is due to the fact that different tasks require different reasoning modules.

\subsection{Generalization for Adaptive Textual Reasoning}

\begin{table}[]
\centering
\caption{Result of the generalization performance in unseen tasks.}
\smalls
\setlength{\tabcolsep}{5pt} 
\renewcommand{\arraystretch}{1.1} 
\begin{tabular}{l|cccccc}
\hline
Env                   & Task & FastWAM & MM-ACT &\begin{tabular}[c]{@{}l@{}}AdaWAM\\ w/o. T.R.\end{tabular} & \begin{tabular}[c]{@{}l@{}}AdaWAM\\ w/o. V.R.\end{tabular} & AdaWAM \\ \hline
\multirow{2}{*}{Seen} & \textit{soup\&cheese}   & 98 & 84 & 96 & 95 & 98 \\
                      & \textit{cheese\&butter} & 93 & 80 & 94 & 92 & 96 \\ \hline
UnSeen                &\textit{ soup\&butter}   & 0  & 24 & 3  & 38 & \textbf{61} \\ \hline
\end{tabular}
\label{tab:generailization}
\vspace{-10pt}
\end{table}

Furthermore, we also find that our model shows generalization in out-of-domain environments with unseen subtask combinations. Specifically, we test the generalization in the LIBERO benchmark by reorganizing the seen task \textit{"put both the alphabet soup and the cream cheese box in the basket"(soup\&cheese)} and \textit{"put both the cream cheese box and the butter in the basket"(cheese\&butter)} to the unseen subtask combination \textit{"put both the alphabet soup and the butter in the basket"(soup\&butter)} and evaluate the success rate in 100 runs. The result is Table~\ref{tab:generailization} indicates that model with no explicit text instruction like FastWAM and AdaWAM w/o. T.R. is unable to generalize to unseen tasks while the CoT textual reasoning benefits the generalization ability for MM-ACT, AdaWAM w/o. V.R. and AdaWAM.

\section{Conclusion}
In this work, we introduced \textbf{AdaWAM}, a World Action Model with adaptive multimodal reasoning whic employs a lightweight dynamic router that autonomously interleaves different reasoning modes based on the execution context. Extensive evaluations across simulated benchmarks (LIBERO, RoboTwin 2.0) and real-world platforms demonstrate that AdaWAM significantly outperforms state-of-the-art embodied policies on long-horizon and complex tasks while reducing unnecessary inference overhead. Moving forward, we plan to extend AdaWAM to additional sensory modalities, such as tactile or depth information, and we will explore unsupervised or reinforcement learning approaches for the routing mechanism, enabling the agent to autonomously learn optimal reasoning boundaries without relying on trajectory-guided annotations.

\section{Limitations}
While AdaWAM significantly advances adaptive multi-modal reasoning and computational efficiency for robotic manipulation, certain aspects remain open for improvement. The current visual observation space is inherently constrained to RGB images, which may limit the spatial reasoning capabilities in geometrically complex or heavily occluded contact-rich tasks. Secondly, the current dynamic routing mechanism relies on supervised learning derived from heuristically annotated data which can be extended to reinforcement learning to enable the agent to autonomously optimize its reasoning pathways and routing decisions based on environmental reward signals, thereby achieving a truly self-adaptive computational paradigm.

\clearpage


\bibliography{example}  

@article{ye2026world,
  title        = {World Action Models are Zero-shot Policies},
  author       = {Ye, Seonghyeon and Ge, Yunhao and Zheng, Kaiyuan and Gao, Shenyuan and Yu, Sihyun and Kurian, George and Indupuru, Suneel and Tan, You Liang and Zhu, Chuning and Xiang, Jiannan and Malik, Ayaan and Lee, Kyungmin and Liang, William and Ranawaka, Nadun and Gu, Jiasheng and Xu, Yinzhen and Wang, Guanzhi and Hu, Fengyuan and Narayan, Avnish and Bjorck, Johan and Wang, Jing and Kim, Gwanghyun and Niu, Dantong and Zheng, Ruijie and Xie, Yuqi and Wu, Jimmy and Wang, Qi and Julian, Ryan and Xu, Danfei and Du, Yilun and Chebotar, Yevgen and Reed, Scott and Kautz, Jan and Zhu, Yuke and Fan, Linxi Jim and Jang, Joel},
  journal      = {arXiv preprint arXiv:2602.15922},
  year         = {2026}
}

@article{yuan2026fastwam,
  title        = {Fast-WAM: Do World Action Models Need Test-time Future Imagination?},
  author       = {Yuan, Tianyuan and Dong, Zibin and Liu, Yicheng and Zhao, Hang},
  journal      = {arXiv preprint arXiv:2603.16666},
  year         = {2026}
}

@article{zawalski2024robotic,
  title        = {Robotic Control via Embodied Chain-of-Thought Reasoning},
  author       = {Zawalski, Micha{\l} and Chen, William and Pertsch, Karl and Mees, Oier and Finn, Chelsea and Levine, Sergey},
  journal      = {arXiv preprint arXiv:2407.08693},
  year         = {2024}
}

@inproceedings{zhao2025cotvla,
  title        = {CoT-VLA: Visual Chain-of-Thought Reasoning for Vision-Language-Action Models},
  author       = {Zhao, Qingqing and Lu, Yao and Kim, Moo Jin and Fu, Zipeng and Zhang, Zhuoyang and Wu, Yecheng and Li, Zhaoshuo and Ma, Qianli and Han, Song and Finn, Chelsea and Handa, Ankur and Liu, Ming-Yu and Xiang, Donglai and Wetzstein, Gordon and Lin, Tsung-Yi},
  booktitle    = {Proceedings of the IEEE/CVF Conference on Computer Vision and Pattern Recognition},
  year         = {2025}
}

@article{du2023learning,
  title        = {Learning Universal Policies via Text-Guided Video Generation},
  author       = {Du, Yilun and Yang, Mengjiao and Dai, Bo and Dai, Hanjun and Nachum, Ofir and Tenenbaum, Joshua B. and Schuurmans, Dale and Abbeel, Pieter},
  journal      = {arXiv preprint arXiv:2302.00111},
  year         = {2023}
}

@article{zhou2024robodreamer,
  title        = {RoboDreamer: Learning Compositional World Models for Robot Imagination},
  author       = {Zhou, Siyuan and Du, Yilun and Chen, Jiaben and Li, Yixuan and Yeung, Dit-Yan and Gan, Chuang},
  journal      = {arXiv preprint arXiv:2404.12377},
  year         = {2024}
}

@article{bharadhwaj2024gen2act,
  title        = {Gen2Act: Human Video Generation in Novel Scenarios Enables Generalizable Robot Manipulation},
  author       = {Bharadhwaj, Homanga and Dwibedi, Debidatta and Gupta, Agrim and Tulsiani, Shubham and Doersch, Carl and Xiao, Ted and Shah, Dhruv and Xia, Fei and Sadigh, Dorsa and Kirmani, Sean},
  journal      = {arXiv preprint arXiv:2409.16283},
  year         = {2024}
}

@article{cheang2024gr2,
  title        = {GR-2: A Generative Video-Language-Action Model with Web-Scale Knowledge for Robot Manipulation},
  author       = {Cheang, Chi and Chen, Guangzeng and Jing, Yuxiang and Kong, Tao and Li, Hang and Li, Yifeng and Liu, Minghuan and Wu, Haoran and Xu, Jiafeng and Yang, Yichu and Zhang, Haozhe and Zhu, Meng},
  journal      = {arXiv preprint arXiv:2410.06158},
  year         = {2024}
}

@article{hu2024vpp,
  title        = {Video Prediction Policy: A Generalist Robot Policy with Predictive Visual Representations},
  author       = {Hu, Yafei and Guo, Yihan and Wang, Peng and Chen, Xiaotong and Wang, Yucheng and Zhang, Jiawei and Sreenath, Koushil and Lu, Cewu and Chen, Jianyu},
  journal      = {arXiv preprint arXiv:2412.14803},
  year         = {2024}
}

@article{zhu2025unified,
  title        = {Unified World Models: Coupling Video and Action Diffusion for Pretraining on Large Robotic Datasets},
  author       = {Zhu, Chuning and Yu, Rui and Feng, Siyuan and Burchfiel, Benjamin and Shah, Priya and Gupta, Abhinav},
  journal      = {arXiv preprint arXiv:2504.02792},
  year         = {2025}
}

@article{pai2025mimicvideo,
  title        = {Mimic-Video: Video-Action Models for Generalizable Robot Control Beyond VLAs},
  author       = {Pai, Jascha and Achenbach, Lucas and Montesinos, Vicente and Forrai, Benedek and Mees, Oier and Nava, Emanuele},
  journal      = {arXiv preprint arXiv:2512.15692},
  year         = {2025}
}

@article{kim2026cosmos,
  title        = {Cosmos Policy: Fine-tuning Video Models for Visuomotor Control and Planning},
  author       = {Kim, Moo Jin and Gao, Yuying and Lin, Tsung-Yi and Lin, Yunfan and Ge, Yunhao and Lam, Garrett and Liang, Percy and Song, Shuran and Liu, Ming-Yu and Finn, Chelsea and Gu, Jiayuan},
  journal      = {arXiv preprint arXiv:2601.16163},
  year         = {2026}
}

@article{li2026causal,
  title        = {Causal World Modeling for Robot Control},
  author       = {Li, Lin and Zhang, Qihang and Luo, Yiming and Yang, Shuai and Wang, Ruilin and Han, Fei and Yu, Mingrui and Gao, Zelin and Xue, Nan and Zhu, Xing and Shen, Yujun and Xu, Yixiao},
  journal      = {arXiv preprint arXiv:2601.21998},
  year         = {2026}
}

@inproceedings{brohan2022rt1,
  title        = {RT-1: Robotics Transformer for Real-World Control at Scale},
  author       = {Brohan, Anthony and Brown, Noah and Carbajal, Justice and Chebotar, Yevgen and Dabis, Joseph and Finn, Chelsea and Gopalakrishnan, Keerthana and Hausman, Karol and Herzog, Alex and Hsu, Jasmine and Ibarz, Julian and Ichter, Brian and Irpan, Alex and Jackson, Tomas and Jesmonth, Sally and Joshi, Nikhil J. and Julian, Ryan and Kalashnikov, Dmitry and Kuang, Yuheng and Leal, Isabel and Lee, Kuang-Huei and Levine, Sergey and Lu, Yao and Malla, Utsav and Manjunath, Deeksha and Mordatch, Igor and Nachum, Ofir and Parada, Carolina and Peralta, Jodilyn and Perez, Emily and Pertsch, Karl and Quiambao, Jornell and Rao, Kanishka and Ryoo, Michael and Salazar, Grecia and Sanketi, Pannag and Sayed, Kevin and Singh, Jaspiar and Sontakke, Sumedh and Stone, Austin and Tan, Clayton and Tran, Huong and Vanhoucke, Vincent and Vuong, Quan and Xia, Fei and Xiao, Ted and Xu, Peng and Xu, Sichun and Yu, Tianhe and Zitkovich, Brianna},
  booktitle    = {Robotics: Science and Systems},
  year         = {2023}
}

@inproceedings{brohan2023rt2,
  title        = {RT-2: Vision-Language-Action Models Transfer Web Knowledge to Robotic Control},
  author       = {Brohan, Anthony and Brown, Noah and Carbajal, Justice and Chebotar, Yevgen and Chen, Xi and Choromanski, Krzysztof and Ding, Tianli and Driess, Danny and Dubey, Avinava and Finn, Chelsea and Florence, Pete and Fu, Chuyuan and Gopalakrishnan, Keerthana and Hausman, Karol and Herzog, Alex and Hsu, Jasmine and Ichter, Brian and Irpan, Alex and Joshi, Nikhil and Julian, Ryan and Kalashnikov, Dmitry and Leal, Isabel and Lee, Kuang-Huei and Levine, Sergey and Lu, Yao and Michalewski, Henryk and Mordatch, Igor and Pertsch, Karl and Rao, Kanishka and Reymann, Karol and Ryoo, Michael and Salazar, Grecia and Sanketi, Pannag and Sermanet, Pierre and Singh, Jaspiar and Sontakke, Sumedh and Stone, Austin and Tan, Clayton and Tran, Huong and Vanhoucke, Vincent and Vega, Steve and Vuong, Quan and Xia, Fei and Xiao, Ted and Xu, Peng and Xu, Sichun and Yu, Tianhe and Zitkovich, Brianna},
  booktitle    = {Proceedings of the 7th Conference on Robot Learning},
  year         = {2023}
}

@article{black2024pi0,
  title        = {$\pi_0$: A Vision-Language-Action Flow Model for General Robot Control},
  author       = {Black, Kevin and Brown, Noah and Driess, Danny and Esmail, Adnan and Equi, Michael and Finn, Chelsea and Fusai, Niccolo and Groom, Lachy and Hausman, Karol and Ichter, Brian and Jakubczak, Szymon and Jones, Tim and Ke, Liyiming and Levine, Sergey and Li-Bell, Adrian and Mothukuri, Mohith and Nair, Suraj and Pertsch, Karl and Shi, Lucy Xiaoyang and Tanner, James and Vuong, Quan and Walling, Anna and Wang, Haohuan and Zhilinsky, Ury},
  journal      = {arXiv preprint arXiv:2410.24164},
  year         = {2024}
}

@article{liu2024rdt,
  title        = {RDT-1B: A Diffusion Foundation Model for Bimanual Manipulation},
  author       = {Liu, Songming and Wu, Lingxuan and Li, Bangguo and Tan, Hengkai and Chen, Huayu and Wang, Zhengyi and Xu, Ke and Su, Hang and Zhu, Jun},
  journal      = {arXiv preprint arXiv:2410.07864},
  year         = {2024}
}

@article{bjorck2025gr00t,
  title        = {GR00T N1: An Open Foundation Model for Generalist Humanoid Robots},
  author       = {Bjorck, Johan and Castaneda, Fernando and Cherniadev, Nikita and Da, Xingye and Ding, Runyu and Fan, Linxi and Fang, Yu and Fox, Dieter and Hu, Fengyuan and Huang, Spencer and Jang, Joel and Jiang, Zhenyu and Kautz, Jan and Kundalia, Kaushil and Lao, Lawrence and Li, Zhiqi and Lin, Zongyu and Lin, Kevin and Liu, Guilin and Llontop, Edith and Magne, Loic and Mandlekar, Ajay and Narayan, Avnish and Nasiriany, Soroush and Reed, Scott and Tan, You Liang and Wang, Guanzhi and Wang, Jing and Wang, Qi and Xiang, Jiannan and Xie, Yuqi and Xu, Yinzhen and Xu, Zhenjia and Ye, Seonghyeon and Yu, Zhiding and Zhang, Ao and Zhang, Hao and Zhao, Yizhou and Zheng, Ruijie and Zhu, Yuke},
  journal      = {arXiv preprint arXiv:2503.14734},
  year         = {2025}
}

@article{gemini2025robotics,
  title        = {Gemini Robotics: Bringing AI into the Physical World},
  author       = {{Gemini Robotics Team} and Abeyruwan, Saminda and Ainslie, Joshua and Alayrac, Jean-Baptiste and others},
  journal      = {arXiv preprint arXiv:2503.20020},
  year         = {2025}
}

@article{zawalski2024ecot,
  title        = {Robotic Control via Embodied Chain-of-Thought Reasoning},
  author       = {Zawalski, Michal and Chen, William and Pertsch, Karl and Mees, Oier and Finn, Chelsea and Levine, Sergey},
  journal      = {arXiv preprint arXiv:2407.08693},
  year         = {2024}
}

@inproceedings{zitkovich2023rt2,
  title        = {RT-2: Vision-Language-Action Models Transfer Web Knowledge to Robotic Control},
  author       = {Zitkovich, Brianna and Yu, Tianhe and Xu, Sichun and Xu, Peng and Xiao, Ted and Xia, Fei and Wu, Jialin and Wohlhart, Paul and Welker, Stefan and Wahid, Ayzaan and Vuong, Quan and Vanhoucke, Vincent and Tran, Huong and Soricut, Radu and Singh, Anikait and Singh, Jaspiar and Sermanet, Pierre and Sanketi, Pannag R. and Salazar, Grecia and Ryoo, Michael S. and Reymann, Krista and Rao, Kanishka and Pertsch, Karl and Mordatch, Igor and Michalewski, Henryk and Lu, Yao and Levine, Sergey and Lee, Lisa and Lee, Tsang-Wei Edward and Leal, Isabel and Kuang, Yuheng and Kalashnikov, Dmitry and Julian, Ryan and Joshi, Nikhil J. and Irpan, Alex and Ichter, Brian and Hsu, Jasmine and Herzog, Alexander and Hausman, Karol and Gopalakrishnan, Keerthana and Fu, Chuyuan and Florence, Pete and Finn, Chelsea and Dubey, Kumar Avinava and Driess, Danny and Ding, Tianli and Choromanski, Krzysztof Marcin and Chen, Xi and Chebotar, Yevgen and Carbajal, Justice and Brown, Noah and Brohan, Anthony and Arenas, Montserrat Gonzalez and Han, Kehang},
  booktitle    = {Proceedings of The 7th Conference on Robot Learning},
  pages        = {2165--2183},
  year         = {2023},
  volume       = {229},
  series       = {Proceedings of Machine Learning Research},
  publisher    = {PMLR}
}

@inproceedings{kim2025openvla,
  title        = {OpenVLA: An Open-Source Vision-Language-Action Model},
  author       = {Kim, Moo Jin and Pertsch, Karl and Karamcheti, Siddharth and Xiao, Ted and Balakrishna, Ashwin and Nair, Suraj and Rafailov, Rafael and Foster, Ethan P. and Sanketi, Pannag R. and Vuong, Quan and Kollar, Thomas and Burchfiel, Benjamin and Tedrake, Russ and Sadigh, Dorsa and Levine, Sergey and Liang, Percy and Finn, Chelsea},
  booktitle    = {Proceedings of The 8th Conference on Robot Learning},
  pages        = {2679--2713},
  year         = {2025},
  volume       = {270},
  series       = {Proceedings of Machine Learning Research},
  publisher    = {PMLR}
}

@inproceedings{ghosh2024octo,
  title        = {Octo: An Open-Source Generalist Robot Policy},
  author       = {Ghosh, Dibya and Walke, Homer Rich and Pertsch, Karl and Black, Kevin and Mees, Oier and Dasari, Sudeep and Hejna, Joey and Kreiman, Tobias and Xu, Charles and Luo, Jianlan and Tan, You Liang and Chen, Lawrence Yunliang and Vuong, Quan and Xiao, Ted and Sanketi, Pannag R. and Sadigh, Dorsa and Finn, Chelsea and Levine, Sergey},
  booktitle    = {Robotics: Science and Systems},
  year         = {2024}
}

@article{Qwen3-VL,
  title={Qwen3-VL Technical Report},
  author={Shuai Bai and Yuxuan Cai and Ruizhe Chen and Keqin Chen and Xionghui Chen and Zesen Cheng and Lianghao Deng and Wei Ding and Chang Gao and Chunjiang Ge and Wenbin Ge and Zhifang Guo and Qidong Huang and Jie Huang and Fei Huang and Binyuan Hui and Shutong Jiang and Zhaohai Li and Mingsheng Li and Mei Li and Kaixin Li and Zicheng Lin and Junyang Lin and Xuejing Liu and Jiawei Liu and Chenglong Liu and Yang Liu and Dayiheng Liu and Shixuan Liu and Dunjie Lu and Ruilin Luo and Chenxu Lv and Rui Men and Lingchen Meng and Xuancheng Ren and Xingzhang Ren and Sibo Song and Yuchong Sun and Jun Tang and Jianhong Tu and Jianqiang Wan and Peng Wang and Pengfei Wang and Qiuyue Wang and Yuxuan Wang and Tianbao Xie and Yiheng Xu and Haiyang Xu and Jin Xu and Zhibo Yang and Mingkun Yang and Jianxin Yang and An Yang and Bowen Yu and Fei Zhang and Hang Zhang and Xi Zhang and Bo Zheng and Humen Zhong and Jingren Zhou and Fan Zhou and Jing Zhou and Yuanzhi Zhu and Ke Zhu},
  journal={arXiv preprint arXiv:2511.21631},
  year={2025}
}

@article{liu2023libero,
  title={LIBERO: Benchmarking Knowledge Transfer for Lifelong Robot Learning},
  author={Liu, Bo and Zhu, Yifeng and Gao, Chongkai and Feng, Yihao and Liu, Qiang and Zhu, Yuke and Stone, Peter},
  journal={arXiv preprint arXiv:2306.03310},
  year={2023}
}

@article{chen2025robotwin,
  title={RoboTwin 2.0: A Scalable Data Generator and Benchmark with Strong Domain Randomization for Robust Bimanual Robotic Manipulation},
  author={Chen, Tianxing and Chen, Zanxin and Chen, Baijun and Cai, Zijian and Liu, Yibin and Liang, Qiwei and Li, Zixuan and Lin, Xianliang and Ge, Yiheng and Gu, Zhenyu and others},
  journal={arXiv preprint arXiv:2506.18088},
  year={2025}
}

@article{wang2026world,
  title={World Action Models: The Next Frontier in Embodied AI},
  author={Wang, Siyin and Shi, Junhao and Fu, Zhaoyang and He, Xinzhe and Liu, Feihong and Yang, Chenchen and Zhou, Yikang and Fei, Zhaoye and Gong, Jingjing and Fu, Jinlan and others},
  journal={arXiv preprint arXiv:2605.12090},
  year={2026}
}

@article{zhang2025pure,
  title={Pure vision language action (vla) models: A comprehensive survey},
  author={Zhang, Dapeng and Sun, Jing and Hu, Chenghui and Wu, Xiaoyan and Yuan, Zhenlong and Zhou, Rui and Shen, Fei and Zhou, Qingguo},
  journal={arXiv preprint arXiv:2509.19012},
  year={2025}
}

@article{ma2026survey,
  title={A Survey on Vision--Language--Action Models for Embodied AI},
  author={Ma, Yueen and Song, Zixing and Zhuang, Yuzheng and Hao, Jianye and King, Irwin},
  journal={IEEE Transactions on Neural Networks and Learning Systems},
  year={2026},
  publisher={IEEE}
}

@article{sapkota2025vision,
  title={Vision-language-action models: Concepts, progress, applications and challenges},
  author={Sapkota, Ranjan and Cao, Yang and Roumeliotis, Konstantinos I and Karkee, Manoj},
  journal={arXiv preprint arXiv:2505.04769},
  year={2025}
}

@article{bi2025motus,
  title={Motus: A unified latent action world model},
  author={Bi, Hongzhe and Tan, Hengkai and Xie, Shenghao and Wang, Zeyuan and Huang, Shuhe and Liu, Haitian and Zhao, Ruowen and Feng, Yao and Xiang, Chendong and Rong, Yinze and others},
  journal={arXiv preprint arXiv:2512.13030},
  year={2025}
}

@article{intelligence2025pi_,
  title={$\pi_{0.5}$: a Vision-Language-Action Model with Open-World Generalization},
  author={Intelligence, Physical and Black, Kevin and Brown, Noah and Darpinian, James and Dhabalia, Karan and Driess, Danny and Esmail, Adnan and Equi, Michael and Finn, Chelsea and Fusai, Niccolo and others},
  journal={arXiv preprint arXiv:2504.16054},
  year={2025}
}

@inproceedings{zhao2025cot,
  title={Cot-vla: Visual chain-of-thought reasoning for vision-language-action models},
  author={Zhao, Qingqing and Lu, Yao and Kim, Moo Jin and Fu, Zipeng and Zhang, Zhuoyang and Wu, Yecheng and Li, Zhaoshuo and Ma, Qianli and Han, Song and Finn, Chelsea and others},
  booktitle={Proceedings of the Computer Vision and Pattern Recognition Conference},
  pages={1702--1713},
  year={2025}
}

@article{zhong2026acot,
  title={ACoT-VLA: Action Chain-of-Thought for Vision-Language-Action Models},
  author={Zhong, Linqing and Liu, Yi and Wei, Yifei and Xiong, Ziyu and Yao, Maoqing and Liu, Si and Ren, Guanghui},
  journal={arXiv preprint arXiv:2601.11404},
  year={2026}
}

@article{liang2025mm,
  title={MM-ACT: Learn from Multimodal Parallel Generation to Act},
  author={Liang, Haotian and Chen, Xinyi and Wang, Bin and Chen, Mingkang and Liu, Yitian and Zhang, Yuhao and Chen, Zanxin and Yang, Tianshuo and Chen, Yilun and Pang, Jiangmiao and others},
  journal={arXiv preprint arXiv:2512.00975},
  year={2025}
}

@article{zheng2025x,
  title={X-vla: Soft-prompted transformer as scalable cross-embodiment vision-language-action model},
  author={Zheng, Jinliang and Li, Jianxiong and Wang, Zhihao and Liu, Dongxiu and Kang, Xirui and Feng, Yuchun and Zheng, Yinan and Zou, Jiayin and Chen, Yilun and Zeng, Jia and others},
  journal={arXiv preprint arXiv:2510.10274},
  year={2025}
}

@article{bu2025agibot,
  title={Agibot world colosseo: A large-scale manipulation platform for scalable and intelligent embodied systems},
  author={Bu, Qingwen and Cai, Jisong and Chen, Li and Cui, Xiuqi and Ding, Yan and Feng, Siyuan and Gao, Shenyuan and He, Xindong and Hu, Xuan and Huang, Xu and others},
  journal={arXiv preprint arXiv:2503.06669},
  year={2025}
}

\clearpage
\appendix
\section{Detailed Information of Benchmarks}~\label{sec:benchmark}
\textbf{LIBERO.}~\citep{liu2023libero} LIBERO is a benchmark for long-horizon robotic manipulation that provides 130 manipulation task suites, from simple goal-conditioned tasks to complex, long-horizon sequences requiring compositional skill reuse.
In our experiment, we test the overall performance in all suites and utilize the LIBERO-10 subset as the hard task subset, which includes multi-object interaction and long-horizon operation. 


\textbf{RoboTwin2.0.}~\citep{chen2025robotwin} RoboTwin2.0 is a high-fidelity dual-arm robotic simulation platform. It provides flexible multi-scene configurations, multi-task setups, and 50 task scenarios covering varying levels of difficulty. In this benchmark, we also select 7 long-term and fine-grained tasks as the hard task subset, including stack bowls, scan objects, pick bottles, etc.

\textbf{Real-World Testebed.} For the real-world test platform, we use the AgileX Split-Type ALOHA platform, featuring a master-follower teleoperation architecture with PiPER 6-DoF arms. We evaluate two long-horizon tasks: Clean up Trash and Wipe Table.

\section{Implementation Details}
We adopt pre-trained Wan2.2-5B for the Visual Reasoning Module and a compressed 1B variant (hidden dimension 1024) for the Action Prediction Module, following \cite{yuan2026fastwam}. The Text Reasoning Module utilizes pre-trained Qwen3-VL-4B. The entire framework is implemented in PyTorch and trained on 8 $\times$ NVIDIA A100 (80GB) GPUs via a two-stage schedule. Stage 1 co-trains the generative backbone and router for 50,000 steps (learning rate $3 \times 10^{-5}$, weight decay 0.005). Stage 2 exclusively fine-tunes the text module for 10,000 steps (learning rate $1 \times 10^{-5}$, weight decay 0.01).

\end{document}